Original Research Paper

# Conceptual Software Engineering Applied to Movie Scripts and Stories

**Sabah Al-Fedaghi**

*Department of Computer Engineering, Kuwait University, Kuwait City, Kuwait*

**Abstract:** According to researchers, a large proportion of research software (software on which researchers rely) is fragile and the source of numerous problems that plague computational science. This study introduces another application of software engineering tools, conceptual modeling, which can be applied to other fields of research. One way to strengthen the relationship between software engineering and other fields is to develop a good way to perform conceptual modeling that is capable of addressing the peculiarities of these fields of study. This study concentrates on humanities and social sciences, which are usually considered "softer" and further away from abstractions and (abstract) machines. Specifically, we focus on conceptual modeling as a software engineering tool (e.g., UML) in the area of stories and movie scripts. Researchers in the humanities and social sciences might not use the same degree of formalization that engineers do, but they still find conceptual modeling useful. Current modeling techniques (e.g., UML) fail in this task because they are geared toward the creation of software systems. Similar Conceptual Modeling Language (e.g., ConML) has been proposed with the humanities and social sciences in mind and, as claimed, can be used to model "anything." This study is a venture in this direction, where a software modeling technique, Thinging Machine (TM), is applied to movie scripts and stories. The paper presents a novel approach to developing diagrammatic static/dynamic models of movie scripts and stories. The TM model diagram serves as a neutral and independent representation for narrative discourse and can be used as a communication instrument among participants. It is based on the notion of thing/machine (or thimac). Things and events are defined as what can be created, released, transferred, received, accepted and processed. Machines are what create, process, release, transfer and receive things. The examples presented include examples from Propp's model of fairytales; the railway children and an actual movie script seem to point to the viability of the approach.

## Introduction

A state of affairs is a combination of things [Wittgenstein] or a complex of things or represents a situation (e.g., a portion of the world). The world is everything that is the case [Wittgenstein]. A thing is "a bit like the way in which primary colors are constitutive of our vision" (Berggruen, 2020). Complexes are similar to thoughts and thus have conceptuality regardless of whether things are thought of as "Being there" or not. However, conceptuality is different from thoughts because they are strictly private, whereas conceptuality is (relative to the representation) public and commonly conceivable. An expression of conceptuality is a conceptual model of a complex state of affairs. A conceptual model can be considered a linguistic expression of a state of affairs conveyed in a diagrammatic form. In software engineering, things must now be thought of in terms of conceptuality (Fayad and Altman, 2001).

According to (Koch, 2020), although conceptual modeling has arguably been practiced over the years, it was not until recently that conceptual engineering became an active topic of research. In this study, the focus is on conceptual modeling as a tool in software engineering, where a certain type of conceptual modeling (e.g., UML) has flourished. The field can be called "conceptual software engineering design" (Simons and Parmee, 2009) or "conceptual software engineering" (SimSE, 2010). Ram *et al.* (2020) reported that some universities offer curricula in conceptual software engineering.

Currently, all areas of academic research (e.g., physics, psychology) are impacted by the method of creating and processing software models (Ram *et al.*, 2020). However, according to (Carver *et al.*, 2018), a large proportion of research software (software on which researchers rely) is developed in an ad hoc manner. As a result, it is fragile and the source of numerous problems that plague modern computational science. Conceptual modeling can not only assist with software but can also be a useful technique for documenting, understanding and communicating artifacts of many kinds. One approach for avoiding many difficulties is to develop a good way to perform modeling tasks capable of addressing the peculiarities of these fields of study. For example, according to (Gonzalez-Pérez, 2020), humanities and social sciences are usually considered "softer" and further away from aspects involving abstractions and (abstract) machines. Gonzalez-Pérez (2020) observed that "people in the humanities and social sciences do a lot of conceptual modeling. They may not use the same terminology and degree of formalization that engineers do, but they still find conceptual modeling useful". It would be helpful to incorporate a suitable conceptual modeling apparatus in these fields of study.

Modeling techniques "imported" straight from engineering (e.g., UML) fail in this task because they are geared toward the creation of software systems (Hug and Gonzalez-Perez, 2020). Hug and Gonzalez-Perez (2012; Gonzalez-Pérez, 2020) proposed a Conceptual Modeling Language (ConML) constructed with the humanities and social sciences in mind that "can be used to model anything". ConML involves class, attribute and association, as do object-oriented modeling languages.

This study is a venture in this direction, where a software modeling technique, Thinging Machine (TM) (Al-Fedaghi, 2020a-d; Al-Fedaghi and Haidar, 2020; Al-Fedaghi and Behbehani, 2020), is applied to movie scripts and stories. Before presenting such applications, we look at the subject of stories and scripts, followed by, for the sake of a self-contained paper, a brief description of TM. Then, we apply the TM modeling to three stories and scripts:

- Narratology fairytale (a type of story)
- Nesbit's children's novel the railway children
- A modern movie script

## Glimpse on Scripts and Stories

Humans, dating back to cavemen (e.g., 30,000-year-old cave drawings in France), have an inclination toward visual representations (Miyagawa *et al.*, 2018). Storytelling has been utilized in various fields of study as a means of conveying information, feelings, ideas, cultural values and human experiences. Throughout history, recorded storytelling technology has proceeded from carvings in wood, impressions in clay tablets and markings in stone and on paper to sophisticated means of recording on film and in digital form. Digital platforms have converged with computerization to provide innovative environments for recording and expressing stories (Sampson *et al.*, 2018). A story can be represented in diverse forms, including narratives, movie scripts and comic books.

Central features of storytelling include plot, characters and setting. According to (Gershon and Page, 2001), such a representation involves dividing the script into two parts: (1) Describing the actions and whereabouts of the different characters and objects and (2) animating the events. Miller (2011) stated that story and storytelling are types of play. An activity is considered play when the activity is done at a special time and in a special space. "Play presents a model of the past and a model for the future" (Miller, 2011). Storytelling is the expression of events in some type of representation, often utilizing certain narrative techniques that supplement elements of exaggeration and emotion. According to (Kosara and Mackinlay, 2013), a story is a sequence of steps in which each step can contain text, images, visualizations, or any combination thereof. Conceptualization of stories captures reality, seeking to release their content in an expanded universe of meanings and interpretations. According to (Kosara and Mackinlay, 2013), a story is "an ordered sequence of steps, with a clearly defined path through it. Each step can contain text, images, visualizations, video, or any combination thereof".

Current diagrammatic methods of transforming representation from text narrative to a diagrammatic form are not straightforward. For example, (Gershon and Page, 2001) asked how a visual presentation can show that the time is early in the morning: How can it show the passage of hours? Another problem is making the transition between disparate pieces of information appear more continuous (Gershon and Page, 2001). Continuity indicates uninterrupted connection, along with succession. In the production of film and television, a script supervisor is concerned with maintaining continuity across shots and along the production process. In comic books, discontinuity is the lack of logical sequence (Ivic, 1999). A gap or break in the chronological sequence, or a fragmented structure, is undesirable.

In software engineering, a user story is a natural language description often written from the perspective of clients, managers, or development team members.

They assist in communication and help software teams organize their understanding of the system and its context (Ralph, 2015). In agile project management and software development, a user story is a tool to describe the desired functionalities of a system from the user's point of view. The concept of a user story goes back to extreme programming (Ralph, 2015).

The aim of this study is to develop a diagrammatic static/dynamic model, abbreviated as TM, of movie scripts and stories using one-category ontology and five generic actions and by defining events in terms of time. The TM model can serve as a vehicle for illustrating "what happened" in the story without concern for "how it is told" and thus can be used as a communication instrument among participants such as the author, designer, director, actors, technical writers and illustrators.

## TM Modeling

Diagrammatic modeling languages hold great promise for software engineering and are able to depict structural and behavioral specifications. However, diagrammatic languages are still considered "doodles" by many practitioners (Bar-Sinai *et al.*, 2016). Bar-Sinai *et al.* (2016) proposed formally defining the operational semantics of involved language. They demonstrated their approach by creating the operational semantics of a type of UML sequence diagram. We claim that the problem originated with the type of diagramming language used in software engineering, where behavior of a system is not well defined. The TM model uses a diagramming language built on one-category ontology and five actions that, when designated over time, produces a behavioral model in terms of the chronology of events.

The TM model articulates the ontology of the world in terms of an entity called thimac (the first three letters of thing and machine) with double faces of the same coin, or two "Being-ness," as a thing and simultaneously as a machine. The first side of the coin exhibits the characterizations assumed by the thimac, while, on the second side of the coin, operational symptoms (processes) emerge that provide a "force" that goes beyond structures or things to embrace other things in the thimac. A thing is subjected to doing (e.g., a tree is a thing Being planted, cut, etc.) and a machine does (e.g., a tree is a machine that absorbs carbon dioxide and uses sunlight to make oxygen). The thing tree and the machine tree are two faces of the thimac tree. A thing is viewed based on (Heidegger's, 1975) notion of thinging. According to (Bryant, 2011), "A tree is a thing through which sunlight, water, carbon dioxide, minerals in the soil, etc., flow. Through a series of operations, the machine transforms those flows of matter, those other machines that pass through it, into various sorts of cells". A thing is a machine and a machine is a thing. While the thing is "The Being of Being," the machine is "Being of becoming." The machine facilitates movement of things, but, simultaneously, the machine itself is a thing in its other mode of Being.

The simplest type of machine is shown in Fig. 1. The actions in the machine (also called stages) can be described as follows:

**Arrive:**   A thing moves to a new machine
**Accept**:   A thing enters a machine. For simplification, we assume that all arriving things are accepted; hence, we can combine, arrive and accept the thing as the **Receive** stage
**Release**:   A thing is marked as ready to be transferred outside the machine (e.g., in an airport, passengers wait to board after passport clearance)
**Process**:   A thing is changed in form, but no new thing results
**Create**:   A new thing is born in a machine
**Transfer**: A thing is input into or output from a machine

Additionally, the TM model includes storage and triggering (denoted by a dashed arrow in this study's figures), which initiates a flow from one machine to another. Multiple machines can interact with each other through movement of things or triggering stages. Triggering is a transformation from one series of movements to another (e.g., electricity triggers cold air).

## Example of TM Modeling

TM modeling can be applied in a variety of systems to create a representation of a portion of reality. The importance of developing a model in a software development effort cannot be overemphasized. However, "the creation of a usable model is no easy task" (Bullinger and Mitra, 2006). In this section, we demonstrate the TM model by re-describing an example that was developed using user stories and use case diagrams.

User stories are initially captured in diagrammatic form using UML use case notation (Bullinger and Mitra, 2006). Bullinger and Mitra (2006) presented a user story called *Stock Goods* as follows:

1. The vendor delivers goods to the store loading dock
2. Accounting pays the vendor for the delivered goods
3. Accounting increments the inventory of delivered goods by the amount supplied by the vendor
4. Accounting notifies the stock person of the arrival of goods
5. The stock person gets the goods from the loading dock
6. The stock person determines the designated location from the inventory
7. The stock person transports the goods from the loading dock to the designated shelf

Figure 2 shows a sample UML diagram that corresponds with this user story.

Figure 3 shows the TM static model of the user story. First, (see my final paper) the vendor with goods (circle 1) goes to the loading dock (2) to be processed (3). Accordingly, the goods arrive (4) and are processed (5) (e.g., counted). The processing of the goods produces the amount of goods (6). This amount and the current total of goods (8) are added (9) to trigger (10) the updating of the total (11).

A notification of the arrival of goods (14) is sent to the stock person (15), as shown in the upper right corner of Fig. 3. The stock person takes the goods from the loading dock (16 and 17), determines their location (18) and transports (19) the goods to the shelf in that location (20).

This diagrammatic description is static in the sense that it represents a stillness (no time) state of affairs. The relationships reflected by the arrows are "logical," not temporal. For example, the arrival of the vendor occurs "before" the counting of the goods, which is a logical and not temporal "before." Logically, the arrival of the vendor can happen years before the counting of goods and, still, the "before" is valid. The diagram in Fig. 3 lacks a structure that applies "meaningfulness" to its parts. The meaningfulness of a part of a system resides in the isomorphism between the part and the thing it is supposed to represent in reality.

Figure 3 is amenable to decomposition exploration to generate a new structural level that involves multiplicity (subsystems). The *structure* of a particular totality in Fig. 3 is the manner in which it is made by actual static components. The figure needs a structure so that its behavior can be specified. Although the wholeness of Fig. 3 is the same, the figure might have different substructures, depending on how it is divided into parts.

The whole-multiplicity notion forms an assemblage of fragments that evolve to facilitate dynamism. Dynamism refers to ordered temporal events. Accordingly, we partition Fig. 3 to create multiplicity that emerges from the whole. This evolutionary change from the whole of Fig. 3 to its parts is utilized to identify the abstract notion of the system's behavior. Decomposition is necessary because the system illustrated in Fig. 3 is obviously "provoked" behaviorally, piece by piece (sub-diagrams). The original single machine of the system is replaced by an assembled machine that is formed from selected submachines.

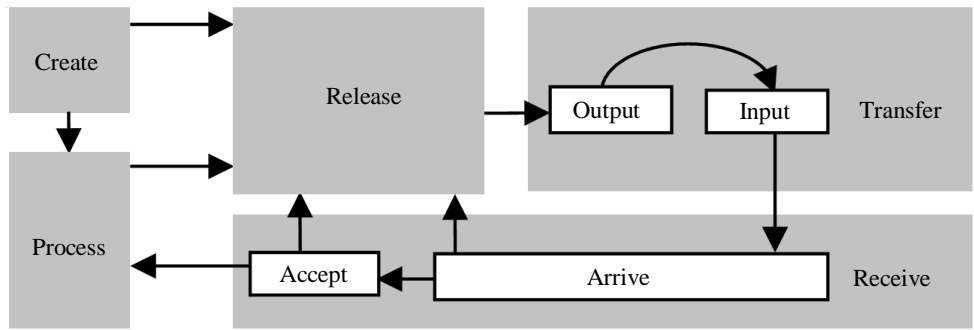

**Fig. 1:** A thinging machine

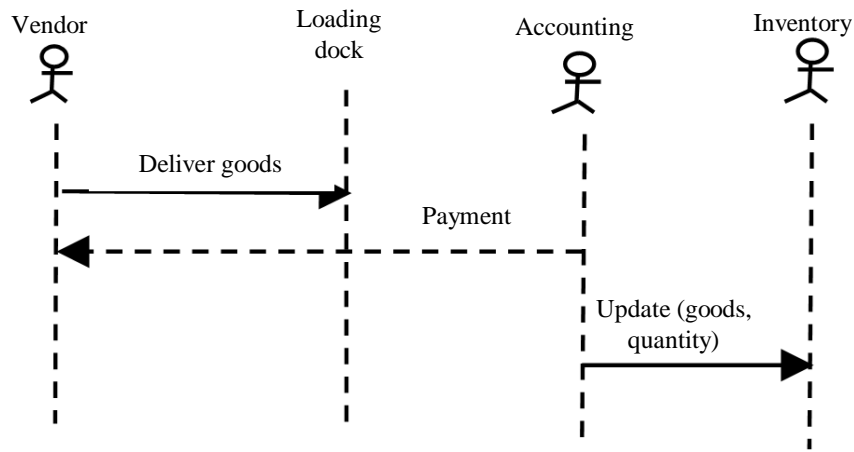

**Fig. 2:** Partial view of a sequence diagram (adopted from Bullinger and Mitra, 2006)

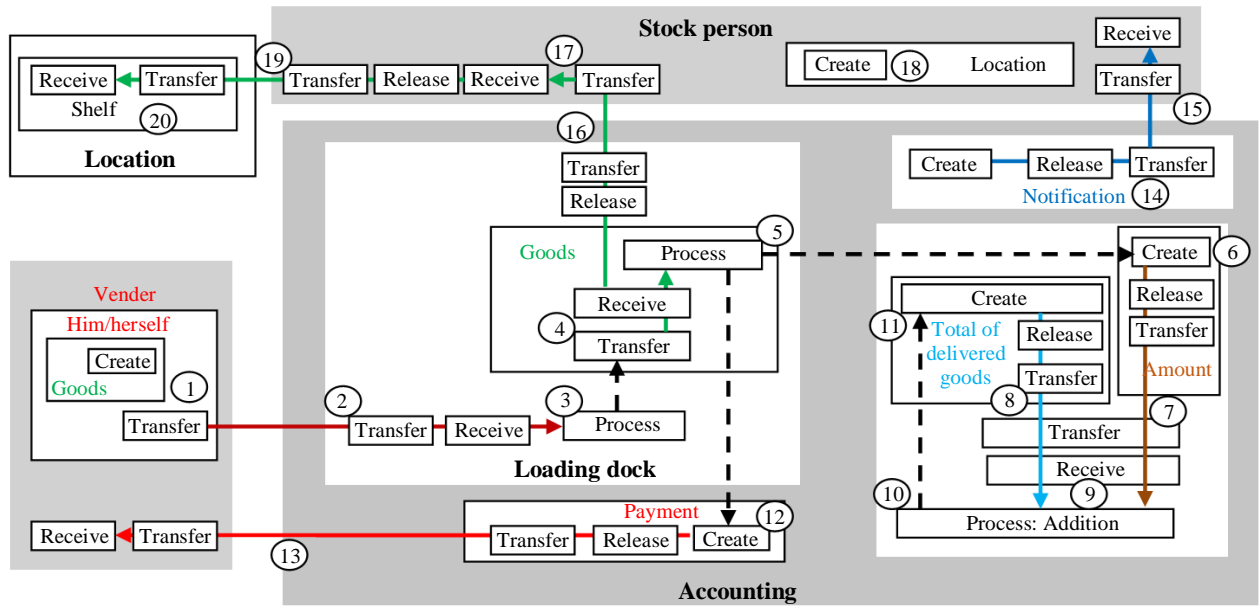

**Fig. 3:** The static TM model of the user story

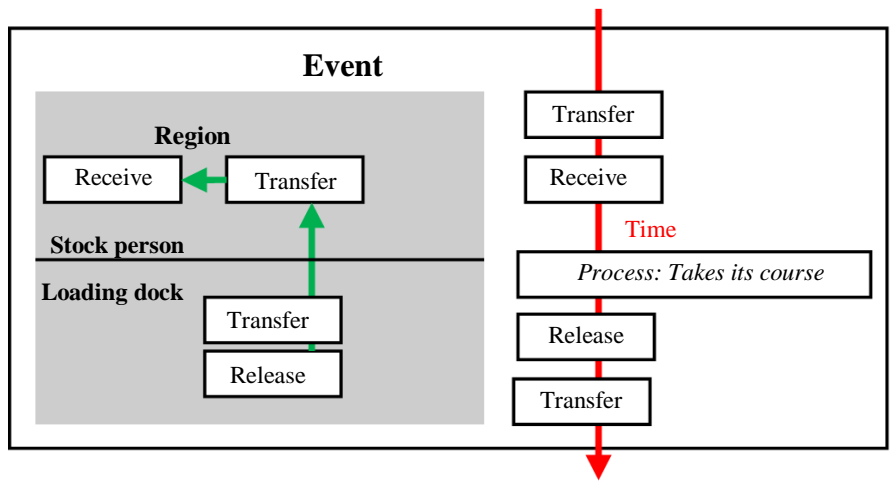

**Fig. 4:** The event the stock person gets the goods from the loading dock

The decomposition of Fig. 3 into sub-diagrams produces events when time is brought into the static picture. Events, not states, are genuine conveyers of behavior. An event is a sub-diagram of Fig. 3 that includes a time machine. For example, Fig. 4 shows the event *The stock person gets the goods from the loading dock.* Note that the region in Fig. 4 is a sub-diagram of Fig. 3.

Accordingly, the following are the events based on a selected decomposition of Fig. 3 and (see Fig. 5):

Event 1 ($E_1$):   The vendor arrives at the store loading dock with goods
Event 2 ($E_2$):   The vendor is processed to deliver the goods
Event 3 ($E_3$):   The goods are received

Event 4 ($E_4$):   The goods are processed to create the amount
Event 5 ($E_5$):   A payment is given to the vendor
Event 6 ($E_6$):   The total amount of goods is updated
Event 7 ($E_7$):   A notification is sent to the stock person
Event 8 ($E_8$):   The stock person gets the goods from the loading dock
Event 9 ($E_9$):   The stock person determines the location of the goods
Event 10 ($E_{10}$):   The stock person transports the goods to the designated shelf

The behavior of the system is given in terms of the chronology of events, as shown in Fig. 6.

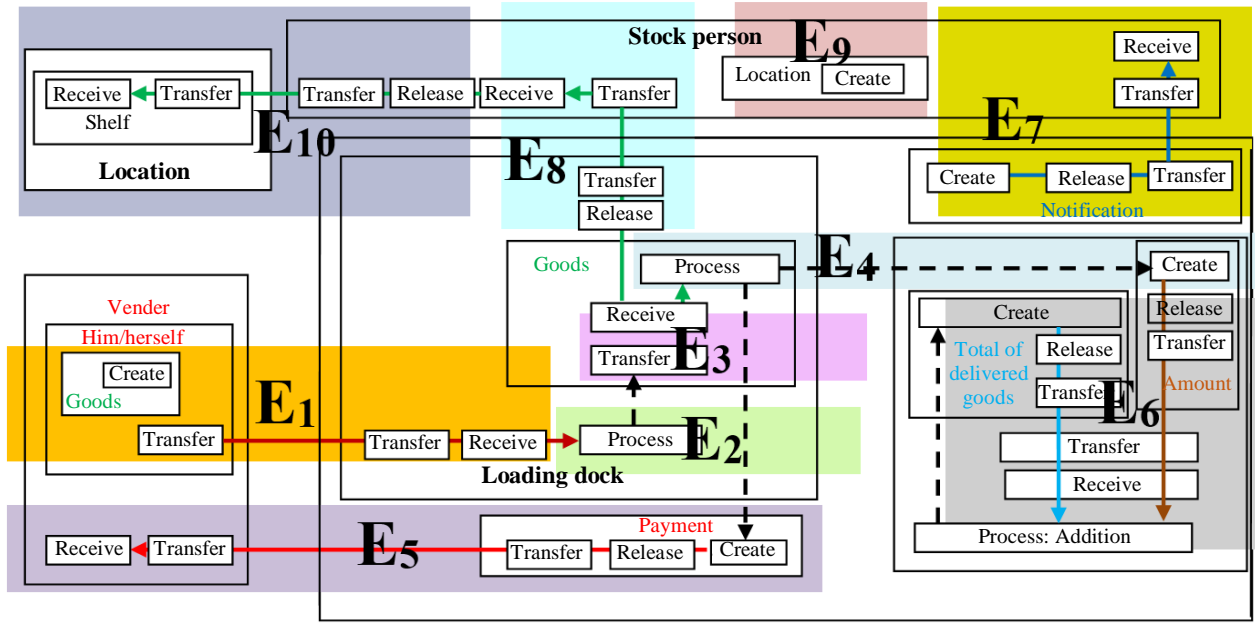

**Fig. 5:** The dynamic TM model of the user story

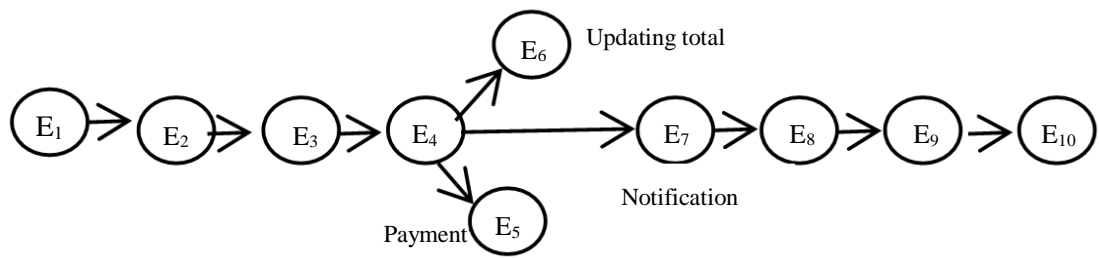

**Fig. 6:** The chronology of events in the user story

The TM model of the stock goods story seems to be less abstracted than the sequence diagram Fig. 2 or other UML diagrams. It has far less notation involving the five basic actions and two types of arrows. The behavioral model can easily be built upon the static description of the story, in comparison with UML diagrams of behavior, such as state or statechart diagrams.

## Propp's Model of Fairytales

Narratology is inspired by (Propp, 1928) classic book on the morphology of the folktale. Propp (1928) structural model of fairytales (a type of story) constructs stories based on a "string" of actions or events called "functions." A function is an action or event (e.g., complicity, struggle, return). A function is defined according to its place in the plot. It is the fundamental unit of analysis (Aguirre, 2011). Tales are organized into sequences; each sequence is composed of functions in temporal order: Function = action (or event) + position in the sequence.

Functions appear in a series. For such series, there is a "move" in the sense of a move in a game such as chess (Aguirre, 2011). Every time a character acts or an event takes place, there is a move-as a part of a series rather than a single action. The term "move" is not quite accurate for such a series. Another rendering might be a "sequence" (Aguirre, 2011). We can observe the informal use of such terms as action, event, function, sequence, series, move and position. Interesting semantics can be drawn when such terms are studied in the context of TM modeling.

The following is a portion of an example of analysis given by (Aguirre, 2011), from (Propp, 1928), with Greek letters denoting functions and sub-functions:

*A man, his wife, two sons, a daughter (α). The brothers, on leaving for work, request their sister to bring lunch to them (β1 γ2); they show the road to the field with shavings (thereby betraying their sister to the dragon*

*δ1). The dragon rearranges the shavings (ε3); the girl goes out to the field with the lunch (δ2) and follows the wrong road (ζ3). The dragon kidnaps her (A1).*

The functions in this portion of the tale are as follows (Aguirre, 2011):

*α* Initial situation (0)
*β* Absentation *One of the members of the family absents himself from home*
*γ* Interdiction *An interdiction is addressed to the hero*
*δ* Violation *The interdiction is violated*
*ε* Reconnaissance *The villain makes an attempt at reconnaissance*
*δ* Delivery *The villain receives information about his victim*
*A* *The villain causes harm or injury to a member of the family*

Figure 7 shows the TM model that corresponds to this narrative. First, there is a man (circle 1 in the figure), wife (2), two sons (3) and a daughter (4). The brothers, on leaving for work (5), ask (6) their sister to bring them lunch (7).

Note how TM expresses the content of the request (8) as a lunch (9) that flows to the brothers at work (10). They show the road (11) to the field with shavings (12), thereby betraying (13) their sister (14) to the dragon (15). The dragon (16) rearranges the shavings (17). The girl goes out to the field (18) with the lunch (19) and follows the wrong road (20). The dragon kidnaps her (21 and 22).

Figure 8 shows the decomposition of Fig. 7 according to the functions in Propp's model. Figure 9 shows the behavioral model in terms of function. We note the string's structural resemblance between the TM model and Propp's model with regard to the decomposition of the story, identifying parts of the story and then associating the parts with functions and TM events Fig. 8. Such an interesting, similar approach to the decomposition of the narrative is worth further investigation. Certainly, this study is not the place to explore such a phenomenon, but this feature facilitates the claim that the TM model seems to be closer to narratology than the diagramming methodology of UML, where different diagrams are developed independently from each other.

## The Railway Children

Nesbit (1906) was an English author and poet who wrote the children's novel the railway children in 1906, which has been adapted for the screen several times. In this section, we apply TM modeling to the following small portion of this novel:

*The children witness a landslide that covers the railway line. The children prevent an imminent accident by waving the girls' red petticoats. The train comes to a stop, just in time, at about twenty metres from where Bobbie stands on the tracks. Weeks later, a ceremony is held at the station to commemorate the children's bravery. The Old Gentleman presents the children with a gold watch each and meets their mother at home. (Nesbit, 1906)*

Figure 10 shows the TM model of the above portion of the novel. A landslide occurs (circle 1) that moves to the railway (2) and accumulates (3 - processed) to cover the railway (4). This scene is witnessed (5) by the children. Note the overlapping among different machines. Even though the landslide is outside the children's machine, it operates simultaneously in their witnessing machine. The waving (7) girls' red petticoats are received (perceived) by the train (8 - the driver). The train is coming from the lift (9) and processing the girls' red petticoat signal (10), which triggers (11) the processing (12 - taking some action, e.g., braking) of the train and triggers stoppage (13) at about twenty meters (14) from where Bobbie stands (15) on the tracks.

The expression "weeks later" mixes the static and dynamic description of the involved activities.

Hence, it is ignored at this point. A ceremony is held at the station (16) attended by the children (17), where the old gentleman (18) presented each of the children (19) with a gold watch (20). The old gentleman also goes to their home (21) to meet (22) the mother (23).

The dynamic model is displayed in Fig. 11. It includes the following events:

Event 1 ($E_1$):    A landslide occurs and covers the railway
Event 2 ($E_2$):    The children witness the landslide
Event 3 ($E_3$):    The train is approaching the landslide area
Event 4 ($E4_3$):   The children wave to the train and Bobbie stands on the tracks
Event 5 ($E_5$):    The train crew take notice of the waving and its meaning.
Event 6 ($E_6$):    The train crew brake the train.
Event 7 ($E_7$):    The train stops about twenty meters (14) from where Bobbie stands (15) on the tracks
Event 8 ($E_8$):    A ceremony is held at the station to commemorate the children's bravery
Event 9 ($E_9$):    The children attend the ceremony
Event 10 ($E_{10}$): The Old Gentleman presents each child with a gold watch
Event 11 ($E_{11}$): The Old Gentleman meets the mother

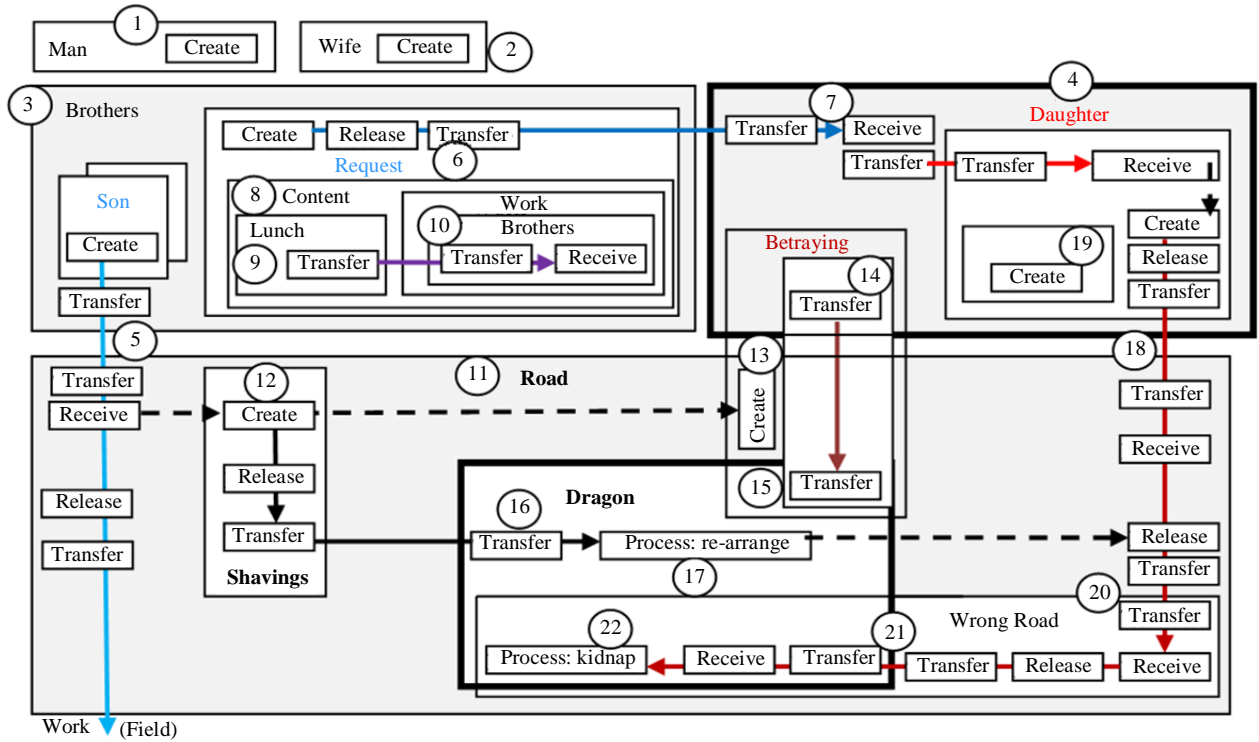

**Fig. 7:** The static TM model of part of the man, wife, sons and daughter scenario

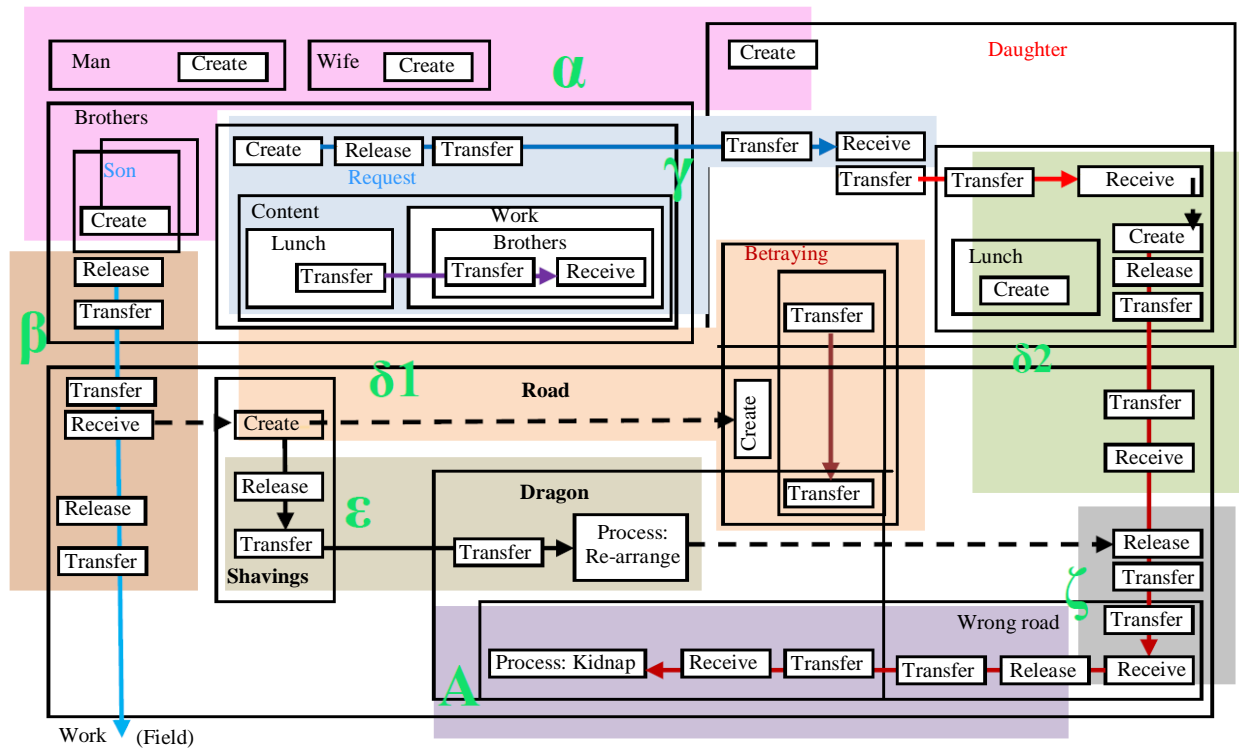

**Fig. 8:** The dynamic TM model of the user story

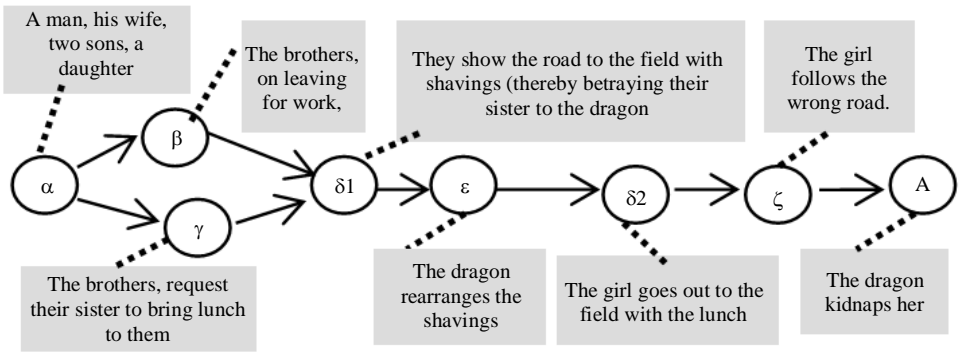

**Fig. 9:** The behavioral TM model of a man, wife, sons and daughter

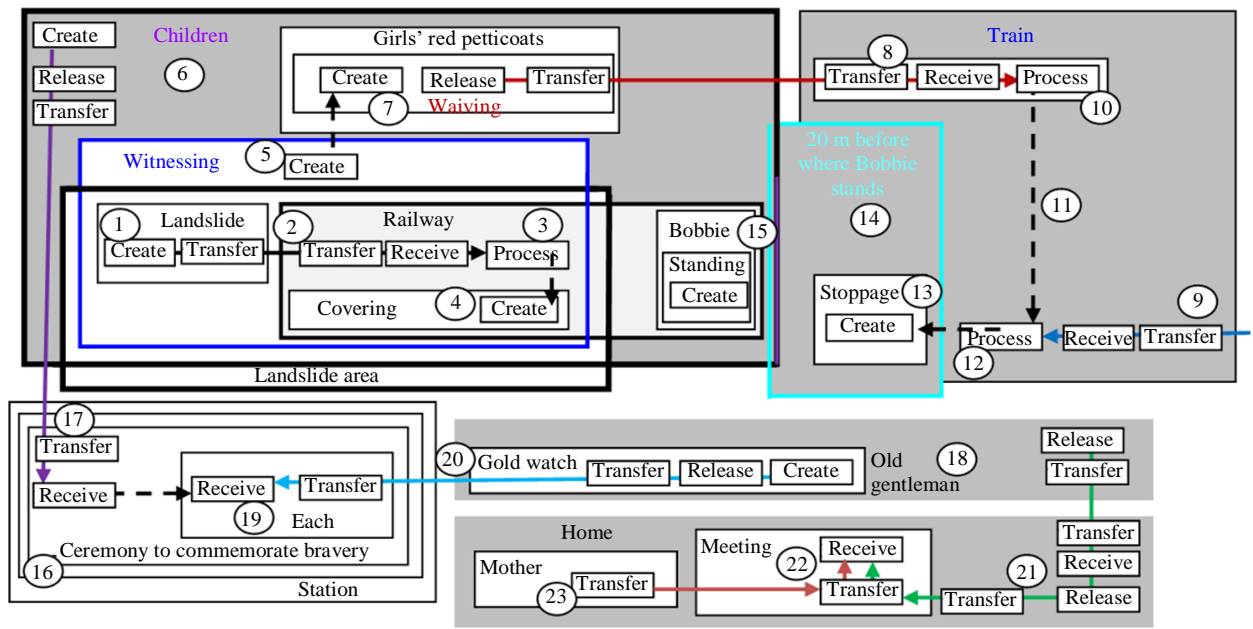

**Fig. 10:** The static TM model of a portion of the railway children

Figure 12 shows the behavioral model of the portion of The Railway Children.

Generally, the TM model seems to express the involved meaning in a reasonable abstract fashion. It seems that it can form a foundation for informal (e.g., oral discussion) and theoretical analysis of the events and their sequences. The novelty of such an application is that the representation is closer to formalized language (mathematics is the ultimate formality), where equivocations and misunderstandings are fewer, but it still represents a medium for communication about the world.

The railway children have been adapted for the screen several times, of which the last was a 1970 film. This brings the issue of how to write a movie script.

## How to Write a Movie Script

According to (Lannom, 2019), understanding how to write a movie script requires knowing the script format and structure. Many scripts begin with scene headings to help "break up physical spaces and give the reader and production team an idea of the story's geography" (Lannom, 2019). Lannom (2019) provided a sample movie script, a portion of which is the following (see the corresponding TM model in Fig. 13):

*We open on a modern suburban home. The front window illuminated by the lights inside. We see the silhouette of a small human figure as it runs back and forth.*

*We push in closer as we slowly see a boy running around the house. A green ball sits on a counter top. A young hand snatches it. It belongs to Fillbert, wiry, lost in his own imaginary world. Dressed as a knight. A toy sword in his other hand.*

*Fillbert: This is my castle. I am sworn to protect it. Anyone that stands in my way shall bear the wrath of the almighty … Just then, the babysitter walks by, trendy, distracted. She is mid-phone call with Fillbert's Mom, Tracy (Lannom, 2019).*

In the diagram in Fig. 13, there is a modern suburban home (1) with a front window (2) that is lit inside (3). The silhouette (4) of a small human figure (5) runs back and forth (6). It is a boy (7) running around the house (8). A green ball (9) sits on a counter top (10). A young hand snatches it (11). It belongs to Fillbert (2), who is wiry and lost in his own imaginary world (13, 14 and 15). He is dressed as a knight (16). He has a toy sword in his other hand (17). For space consideration, we ignore what Fillbert is shouting. The babysitter (18) walks by (19) and she is trendy, distracted (20). She is mid-phone call with Fillbert's mom, Tracy (21, 22 and 23).

The selected events are as follows Fig. 14:

Event 1 ($E_1$): There is a modern suburban home
Event 2 ($E_2$): The front window is illuminated by the lights inside, where we see the silhouette of a small human figure as it runs back and forth
Event 3 ($E_3$): It is a boy running around the house
Event 4 ($E_4$): A green ball sits on a counter top. A young hand snatches it
Event 5 ($E_5$): It belongs to Fillbert, who is wiry and lost in his own imaginary world
Event 6 ($E_6$): He is dressed as a night, a toy sword in his hand
Event 7 ($E_7$): The babysitter walks by.
Event 8 ($E_8$): The babysitter is trendy and distracted
Event 9 ($E_9$): The babysitter is mid-phone call with Fillbert's mom

Figure 15: Shows the behavioral model of the script.

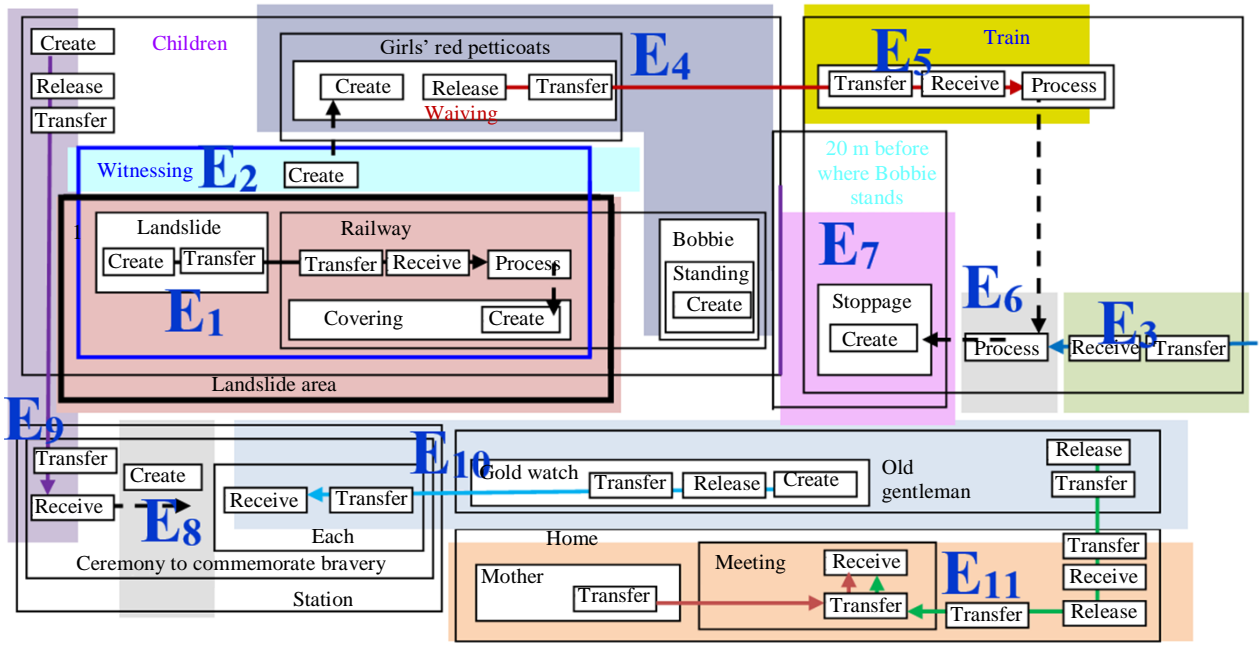

**Fig. 11:** The dynamic TM model of a portion of The Railway Children

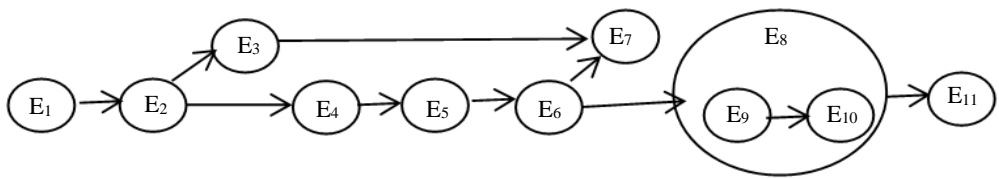

**Fig. 12:** The behavioral model of a portion of the railway children

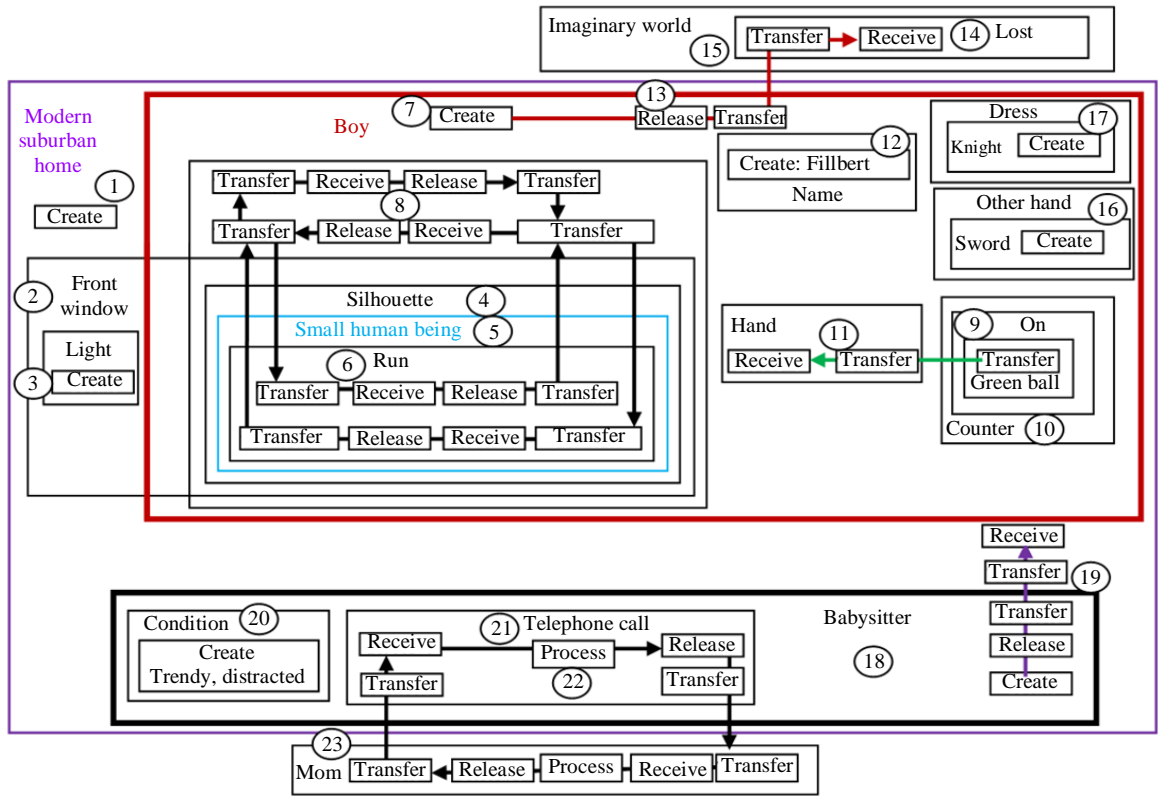

**Fig. 13:** The static model of a portion of the portion of the movie script

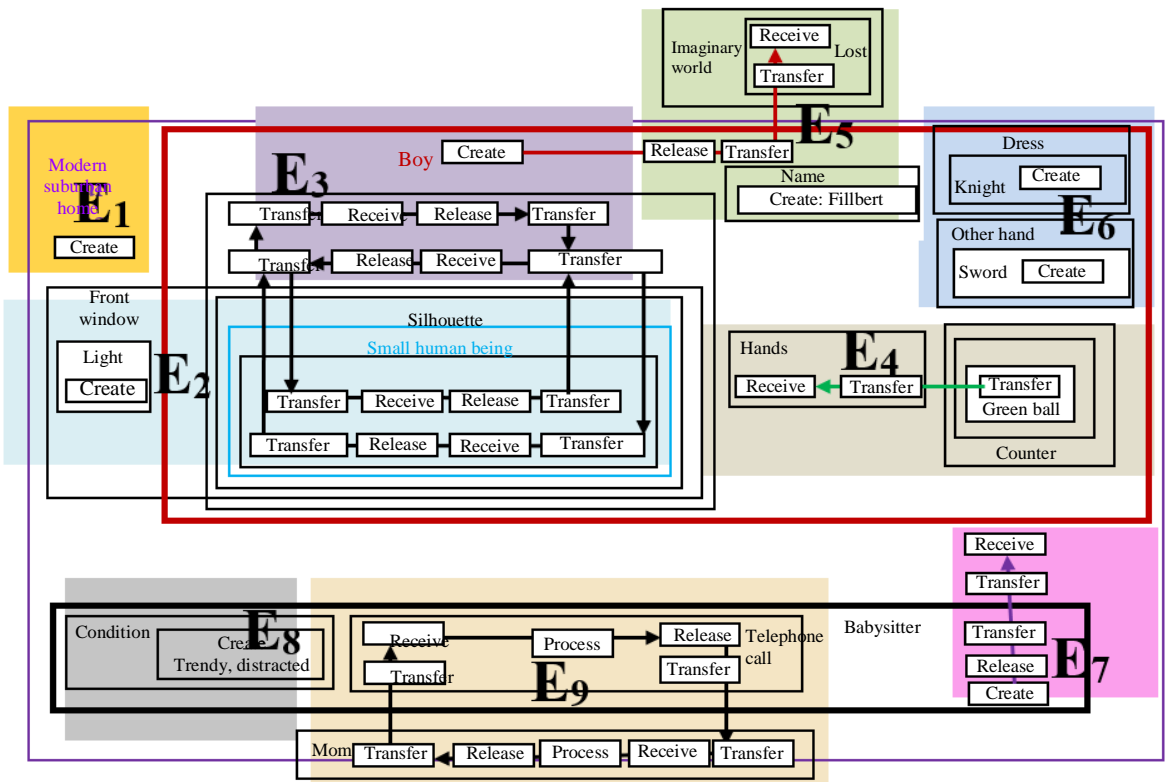

**Fig. 14:** The dynamic model of a portion of the movie script

1728

## Analysis

While the TM modeling has been applied in requirement analysis in software engineering, as demonstrated in (Bullinger and Mitra's, 2006) user story, *stock goods* (section 4: Example of TM Modeling), it is also suitable for diagrammatically representing stories and scripts. Both applications have a narrative form of presentation. The diagrammatic representation is as helpful in unfolding stories and scripts as in describing requirements in software engineering. Such a connection can benefit both fields of study and provide a more engineering-like form for developing stories and scripts.

## Conclusion

In this study, we aimed to develop a diagrammatic static/dynamic modeling of stories and movie scripts Fig. 16. The resultant representation can be utilized as a vehicle for communication among participants, such as author, designer, director, actors, technical writers and illustrators. Additionally, the approach can be used in education where students (e.g., university) can analyze and scrutinize the story or the script.

An immediate aim of applying TM in this field is to explore the plausibility of the idea. The examples discussed here point to the viability of the approach. The TM description can be utilized to supplement text or even replace it. One weakness of the proposed approach is its applicability to long stories and scripts. One solution to this problem is to apply the TM modeling to pieces of the story or scripts. Further experimentation with larger portions of stories and scripts will be conducted in the future to substantiate these claims.

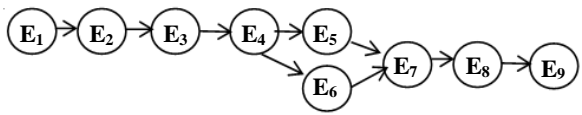

**Fig. 15:** Shows the behavioral model of a portion of the movie script

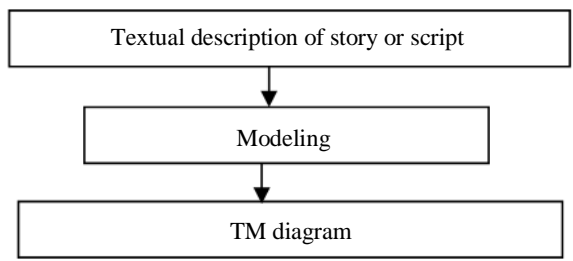

**Fig. 16:** A general view of the involved research problem

## Ethics

This article is original and contains unpublished material. No ethical issues were involved and the authors have no conflicts of interest to disclose.